\documentclass{article}

\usepackage{microtype}
\usepackage{graphicx}
\usepackage{subfigure}
\usepackage{booktabs} 

\usepackage{hyperref}

\usepackage[accepted]{icml2023_dp4ml}

\usepackage{amsmath}
\usepackage{amssymb}
\usepackage{mathtools}
\usepackage{amsthm}

\usepackage[capitalize,noabbrev]{cleveref}

\theoremstyle{plain}

\theoremstyle{definition}

\theoremstyle{remark}

\usepackage[textsize=tiny]{todonotes}

\newtheorem*{prop*}{Proposition}

\theoremstyle{definition}

\DeclareMathOperator{\Tr}{Tr}

\newcommand{\argmin}[1]{\underset{#1}{\operatorname{arg}\,\operatorname{min}}\;}

\newcommand{\x}{{\bf x}}
\newcommand{\y}{{\bf y}}
\newcommand{\z}{{\bf z}}

\newcommand{\M}{{\bf M}}

\newcommand{\R}{\mathbb{R}}

\newcommand{\V}{{\bf V}}
\newcommand{\W}{{\bf W}}
\newcommand{\X}{{\bf X}}
\newcommand{\Y}{{\bf Y}}
\newcommand{\Z}{{\bf Z}}

\newcommand{\Real}{{\mathbb R}}

\icmltitlerunning{Unlocking the Potential of Similarity Matching}

\begin{document}

\twocolumn[
\icmltitle{Unlocking the Potential of Similarity Matching: \\ Scalability, Supervision and Pre-training}
%


\icmlsetsymbol{equal}{*}

\begin{icmlauthorlist}
\icmlauthor{Yanis Bahroun}{equal,yyy}
\icmlauthor{Shagesh Sridharan}{equal,comp}
\icmlauthor{Atithi Acharya}{comp,sch}
\icmlauthor{Dmitri Chklovskii}{yyy}
\icmlauthor{Anirvan M. Sengupta}{comp,sch,mat}
\end{icmlauthorlist}

\icmlaffiliation{yyy}{Center for Computational Neuroscience, Flatiron Institute, New York City, USA}
\icmlaffiliation{comp}{Department of Physics and Astronomy, Rutgers University, New Brunswick, New Jersey, USA}
\icmlaffiliation{sch}{Center for Computational Quantum Physics, Flatiron Institute, New York City, USA}
\icmlaffiliation{mat}{Center for Computational Mathematics, Flatiron Institute, New York City, USA}

\icmlcorrespondingauthor{Yanis Bahroun}{ybahroun@flatironinstitute.org}
\icmlcorrespondingauthor{Shagesh Sridharan}{shagesh.sridharan@rutgers.edu}

\icmlkeywords{Machine Learning, ICML}
\icmlkeywords{Backpropagation, Biological plausibility, Online learning, Local learning, Unsupervised similarity matching, Convolutional nonnegative similarity matching, PyTorch implementation, Localized supervised similarity matching objective, Canonical correlation analysis, Pre-training, LeNet}

\vskip 0.3in
]



\printAffiliationsAndNotice{\icmlEqualContribution} 

\begin{abstract}
While effective, the backpropagation (BP) algorithm exhibits limitations in terms of biological plausibility, computational cost, and suitability for online learning. As a result, there has been a growing interest in developing alternative biologically plausible learning approaches that rely on local learning rules. This study focuses on the primarily unsupervised similarity matching (SM) framework, which aligns with observed mechanisms in biological systems and offers online, localized, and biologically plausible algorithms. i) To scale SM to large datasets, we propose an implementation of Convolutional Nonnegative SM using PyTorch. ii) We introduce a localized supervised SM objective reminiscent of canonical correlation analysis, facilitating stacking SM layers. iii) We leverage the PyTorch implementation for pre-training architectures such as LeNet and compare the evaluation of features against BP-trained models. This work combines biologically plausible algorithms with computational efficiency opening multiple avenues for further explorations.
%
%
%
%
%
%
%
%
\end{abstract}

\section{Introduction} \label{introduction}
%
%
%
%

The backpropagation (BP) algorithm \cite{rumelhart1986learning} has proven highly effective; however, its reliance on computing gradients across all network layers raises concerns regarding its biological plausibility. Additionally, BP necessitates extensive computations and information exchange throughout the network, resulting in significant computational expense and energy consumption. Furthermore, the batch-based nature of BP makes it less suitable for online learning scenarios. Consequently, there has been a surge of interest in developing localized and biologically plausible alternatives to the widely adopted BP algorithm in recent years \cite{sacramento2018dendritic,belilovsky2019greedy,duan2022training,golkar2022constrained}.

%
%
%

%

%
Biological systems rely on localized computations, demonstrate energy efficiency, and can train online. Therefore, developing algorithms that align with these principles is imperative. Such efforts can yield more efficient and scalable learning processes while facilitating continual learning. In the machine learning literature, localized learning is a specific area of research that explicitly addresses the trade-off between model performance, training speed, and scalability. Localized learning encompasses a class of machine learning methods that update model parameters based on local objectives, focusing on optimizing the performance of individual layers at a time, often referred to as greedy optimization.

In this study, we explore the potential of the similarity matching (SM) framework \cite{pehlevan2019neuroscience,lipshutz2023normative} as a promising candidate for localized learning. The SM framework is a normative approach introduced to derive and comprehend the algorithmic basis of neural computation, with a particular emphasis on unsupervised problems. It involves deriving algorithms from computational objectives and evaluating their compatibility with anatomical and physiological observations. The resulting algorithms exhibit desirable properties such as locality, online trainability, and biological plausibility.

Previous studies suggested using nonnegative SM (NSM) for feature learning as individual or stacked layers \cite{bahroun2017online,bahroun2017building,obeid2019structured}. However, these fully unsupervised NSM models demonstrated limited performance when stacked together. To address these limitations, the contrastive NSM (CNSM) was proposed \cite{qin2021contrastive} to alleviate the issues above.
While CNSM employs local learning rules, it lacks layer-wise training. Furthermore, the architecture of the NSM network involves lateral inhibition, which presents challenges in terms of implementation when scaling the model on GPUs and integrating it with convolution. 

Our proposed approach involves refining the implementation of deep NSM and adapting it for large-scale datasets. By leveraging this enhanced deep NSM framework, we aim to uncover its potential for localized learning and address the limitations observed in earlier models. Through empirical evaluations and experiments, we demonstrate the effectiveness and scalability of our proposed methodology, shedding light on the capabilities and advantages of deep NSM for addressing complex learning tasks.

%
%
%

%
%
%
%

Our study makes several key contributions. Firstly, in Section \ref{sec:scaling}, we present the implementation of Convolutional NSM using PyTorch. We employ tensor network methods to compute the neural dynamics and utilize auto differentiation to update the network parameters based on the energy function. This novel implementation enables efficient and scalable computation of Convolutional NSM.

Secondly, in Section \ref{sec:supervision_NSM}, we propose a localized supervised objective for SM, which is trained greedily. This objective resembles canonical correlation analysis and allows us to incorporate supervisory signals at different layers of the network. This model can also be adapted as a supervised learning algorithm, leveraging labeled and unlabeled data to enhance performance.

Finally, in Section \ref{sec:pre-training}, leveraging the benefits of the PyTorch implementation, we propose pre-training architectures such as LeNet. We compare the learned features when the model is trained using BP. This investigation serves as a first step towards NSM-based pre-training of large-scale models, providing insights into the potential benefits and performance of Convolutional NSM compared to BP-trained models.

Overall, our contributions encompass the development of a PyTorch implementation for Convolutional NSM, introducing a localized supervised objective, and exploring pre-training using Convolutional NSM. Through our novel approaches and evaluations, we advance the understanding of localized learning methods and their potential for enhancing model training, feature learning, and pre-training strategies.

\paragraph*{Notations.}
For positive integers $n$, $m$, let $\Real^n$ denote $n$-dimensional Euclidean space, let $\Real^n_+$ denote the $n$-dimensional positive Euclidean space, and let $\Real^{n\times m}$ denote the set of $n \times m$ matrices equipped with the Frobenius norm $\| \cdot \|_F$ and the transpose operator $\cdot^\top$. Boldface lowercase letters (e.g., $\x_t$) denote vectors, and boldface uppercase letters (e.g., $\M$) denote matrices.
For a set of inputs $\x_t \in \Real^n$, $t = 1,\ldots, T$, we denote by $\X \in \Real^{n \times T}$ the input matrix where the columns of the matrix are the input vectors. Similarly, we define the label matrix $\Y \in \Real^{c \times T} = [\y_1,\ldots, \y_T ]$ where $\y_i$ is the one hot encoding of $c$ classes, and the $k^\text{th}$ layer encoding matrix $\Z_k \in \Real^{m_k \times T} = [\z_{k,1},\ldots, \z_{k,T} ]$.

%
%
%
%
%
%


\section{Background and related work}\label{sec:background}

To build biologically plausible neural networks (NNs) from a normative approach, the authors of \cite{pehlevan2014hebbian} considered an objective function reminiscent of classical multidimensional scaling \cite{cox2000multidimensional}. 
They departed from the standard reconstruction approach, which had led to the popular Oja's algorithms \cite{oja1989neural}. 
Indeed, existing models have led to non-local learning rules, which contradict the fundamental Hebbian principle of plasticity, stating that the connections between two neurons are strengthened when activated simultaneously. 
The objective they considered is the following 
\begin{align}\label{eq:NSM_Orig}
   \hat{\Z} = \argmin{\Z \in \Real_{+}^{m\times T} } \| \X^\top \X  - \Z^\top \Z \|_F^2.
\end{align}
%
%
%
While the NSM objective \eqref{eq:NSM_Orig} can be minimized by taking gradient descent steps with respect to $\Z$, this would not lead to an online algorithm because such computation requires combining data from different time steps. 
Rather, the authors of \cite{pehlevan2017similarity} introduced auxiliary matrix variables, $\W$ and $\M$ allowing for the NSM computation using solely contemporary inputs and will correspond to synaptic weights in the network implementation and rewrite the minimization problem \eqref{eq:NSM_Orig} as the following min-max problem
%
%
%
\begin{align}\label{eq:minmax_NSM}
 &\min_{\Z\in\R_{+}^{m\times T},\W}\max_{\M} -4 \Tr(\X^\top \W^\top \Z -\tfrac{1}{2} \Z^\top \M^\top \Z ) \nonumber
    \\ &\qquad \quad + 2 \Tr(\W^\top \W) - \Tr(\M^\top \M). 
\end{align}
The optimal solution of Eq.~\eqref{eq:minmax_NSM}, $\hat{\Z}$ can be obtained by running the following dynamics until convergence
\begin{align}\label{eq:y_dynamics}
    \tfrac{d\Z(\gamma)}{d\gamma} = [\W \X - \M \Z(\gamma)]_+~~, 
\end{align}
%
%
where $[\cdot]_+$ is the point-wise ReLU. The update rules for the parameters $\W, \M$ admit closed-form solutions as follows
\begin{align}\label{eq:wm_update}
    \Delta \W = \X \hat{\Z}^\top ~~,~~ \Delta \M = - \hat{\Z}  \hat{\Z}^\top.
\end{align}
%

\section{Scaling NSM: PyTorch implementation}\label{sec:scaling}

To ensure the scalability of the NSM framework, we leverage the powerful capabilities of the PyTorch library. We can significantly enhance computational performance by utilizing tools that enable efficient GPU utilization. Our implementation includes a versatile pipeline that can accommodate various similarity-matching objective functions, as extensively explored in prior publications \cite{bahroun2017online, sengupta2018manifold}. In particular, our approach is well-suited for loss functions incorporating local error signals, and we effectively approximate the solution using a convolutional architecture. This tailored implementation allows for robust and efficient computations, enabling us to address a wide range of similarity-matching objectives within the NSM framework.

\subsection{Implementation} \label{sec:PyTorch-NSM}

\paragraph{Neural dynamics.}

\begin{figure}[!h]
    \centering
    \includegraphics[width=0.35\textwidth]{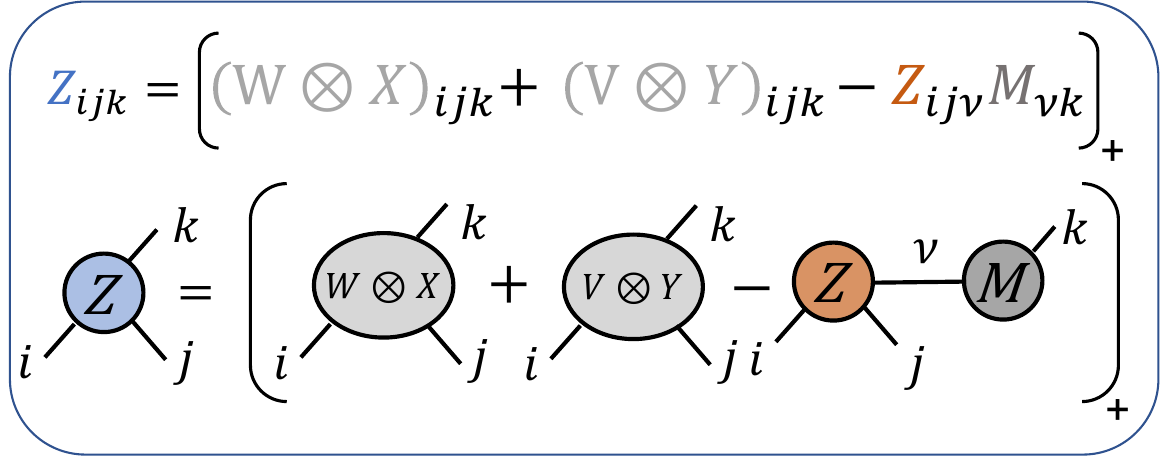}
    \vspace{-0.2cm}
    \caption{ Graphical notation of tensor operations for visualization underlying the efficient computations. 
    Each colored node denotes a tensor, and each outgoing line indicates a dimension of the tensor. 
    We assume implicit summation over indices shared by two tensors, e.g., the index $\nu$ shared by $\Z$ and auxiliary variable $\M$ here. }
    \vspace{-0.4cm}
    \label{fig:contr}
\end{figure}

Each incoming input image, denoted as $\X$, can be represented as a 3-dimensional tensor with dimensions for channel, width, and height. To process each input, we execute the dynamics described in Eq.~\ref{eq:y_dynamics} until convergence. For efficient and repeated calculations and updates of the tensor $\M\Z$, we employ the EinOps library \cite{rogozhnikov2022einops}. EinOps leverages Einstein notations, a mathematical representation for expressing tensor operations, as visualized in Figure \ref{fig:contr}. This notation allows us to match tensor axes based on labels, such as $\nu$ between $\M$ and $\Z$, as depicted in Figure \ref{fig:contr}. By utilizing EinOps within the PyTorch framework, we can preserve the tensor structure, avoid the need to compute permutations of axes and achieve the computational speedup offered by PyTorch.

%

\paragraph{Gradient updates.}
We evaluate the loss function defined in Eq.~\ref{eq:minmax_NSM} using the current values of the auxiliary variables $\W$, $\M$, and the previously computed $\Z$.
Next, we utilize the automatic gradients provided by PyTorch to perform updates to the auxiliary variables according to Eq.~\ref{eq:wm_update}.
%
%


\subsection{Numerical evaluation}

We compare our novel implementation with K-means \cite{lloyd1982least} and Manifold tiling \cite{sengupta2018manifold} techniques using the CIFAR-10 dataset \cite{krizhevsky2009learning}. Our experimental setup follows the methodology outlined in \cite{bahroun2017online}. In our approach, we employ a convolutional NSM, where the input is composed of patches represented as $\X$ in Eq.~\eqref{eq:y_dynamics}. For image classification, we utilize an SVM and employ a simple pooling technique on the feature vectors $\Z$. 

%
%
%
%
%
\begin{figure}[!h]
\begin{center}
    \vspace{-0.1cm}
    \includegraphics[width=.40\textwidth]{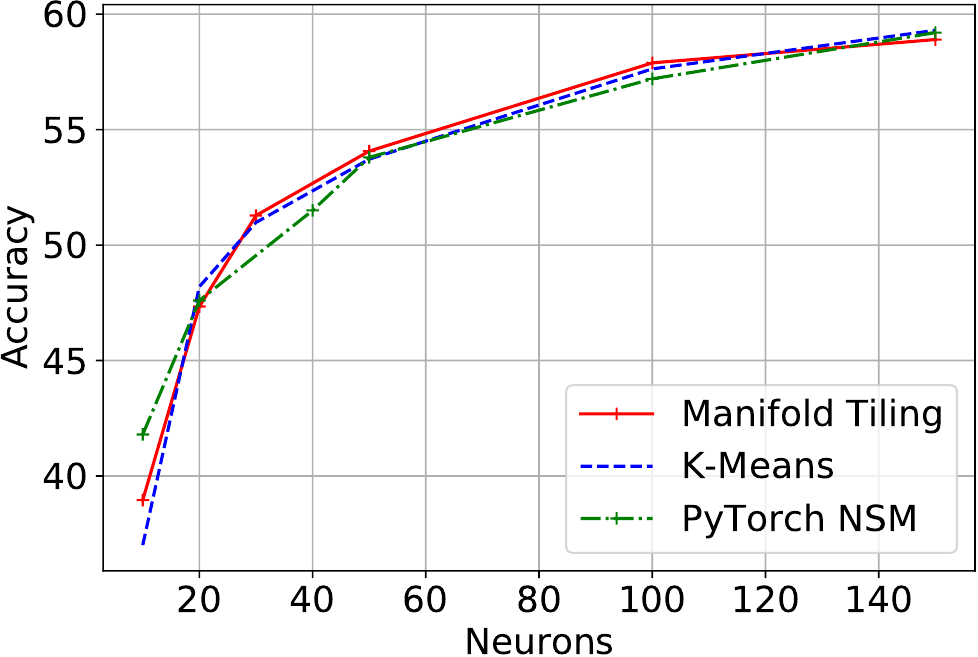}
    \vspace{-0.3cm}
    \caption{Our algorithm achieves the same accuracy as the solution implemented in \cite{bahroun2017online, sengupta2018manifold} despite significantly lower time complexity (c.f. Table~\ref{table:compute_time}).}
    \label{fig:accuracy}
    \end{center}
    \vspace{-0.2cm}
\end{figure}

By leveraging the computational power of GPUs, our single-layer NN achieves significantly faster convergence to a steady state. To provide a quantitative comparison, we present the training times for processing 10,000 images using different algorithms in Table \ref{table:compute_time}. Specifically, we include the computation times for a conventional implementation executed on a CPU, as well as the CPU and GPU implementations of our novel approach. The conventional implementation extracts patches from images and streams patches to the network in an online fashion after initial processing. PyTorch speeds up this computation using convolution layers, and EinOps enables it to happen on GPUs.

%
%
\begin{table}[!h]
\centering
\begin{tabular}{ |c|c|c|c| } 
\hline
 Algorithm  & Conventional & CPU\footnotemark[1] & GPU\footnotemark[2] \\ 
 \hline
 10k images & 2399s & 93.85s & 13.54s \\ 
 \hline
\end{tabular}
\caption{Training times for processing 10,000 images using different algorithms on CPU and GPU.}
\vspace{-.3cm}
\label{table:compute_time}
\end{table}

\footnotetext[1]{CPU: Intel Xeon Platinum 8358 Processors (48MB cache, 2.60GHz): 3200 MHz DDR4 memory.}
\footnotetext[2]{GPU: 1xTesla T4 GPU, 2560 CUDA cores, compute 3.7, 15GB GDDR6 VRAM.} 

The PyTorch and conventional implementations yield similar results, accounting for potential variations due to initialization, as shown in Fig.~\ref{fig:accuracy}. The comparable performance between the two implementations supports the effectiveness and reliability of our novel GPU-accelerated approach, further validating its potential for efficient training of NNs.
%

\section{Including supervision in NSM} \label{sec:supervision_NSM}

In practice, it has been observed that stacking layers of unsupervised learning algorithms offers limited improvements beyond a certain depth \cite{he2014unsupervised,bahroun2017online,amato2019hebbian}.  Recent studies investigating feature alignment in deep NNs shed light on this phenomenon. It has been demonstrated that as the depth of the layer increases, the alignment between the learned features and the input samples decreases, while the alignment with the corresponding labels becomes inversely related \cite{canatar2022kernel}.  
%
In this section, we propose incorporating supervision within the SM framework while preserving the ability to train locally and greedily. By introducing this localized supervision, we aim to address the challenges associated with feature alignment in deep NNs and enhance the performance of the SM framework.

%
%
%
%

\subsection{Supervised objective}
Armed with the novel implementation of NSM, we propose a new type of supervised SM (S$^2$M) defined as follows for $k \in \{1, L\}$ where $L$ is the number of layers,
\begin{align}\label{eq:S2M_obj}
    \hat{\Z}_k = \argmin{\Z_k \geq 0} \left \| \left[\hat{\Z}_{k-1}^\top \hat{\Z}_{k-1} + \alpha_k \Y^\top \Y \right] - \Z_{k}^\top \Z_{k} \right \|_F^2 ,
\vspace{-0.4cm}
\end{align}
with $\Y \in \Real^{c \times T}$, the matrix of labels, with one-hot encoding of $c$ classes and $\alpha_k$ controls the influence of the label matrix. 
For simplicity, we absorb $\alpha_k$ into $\Y^\top \Y$.
The PyTorch implementation of S$^2$M is similar to that of the model defined in the previous section.
We introduce auxiliary variables $\W_k$, $\M_k$, and $\V_k$ and obtain the neural activity using EinOps by running the following dynamics until convergence 
\begin{align}\label{eq:y_dynamics_supervised}
    \tfrac{d\Z_{k}(\gamma)}{d\gamma} = [\W_k \hat{\Z}_{k-1} + \V_k \Y - \M_k \Z_{k}(\gamma)]_+~~. 
\end{align}
The update rules are implemented using auto differentiation of the following min-max objective
\begin{align}
    &\max_{\M_k}\min_{\W_{k},\V_{k}} -4 \Tr(\left[\Z_{k-1}^\top \W_{k}^\top +\Y^\top \V_{k}^\top -\tfrac{1}{2} \Z_{k}^\top \M_{k}^\top\right] \Z_{k}) \nonumber
    \\ &\qquad \quad + 2 \Tr(\W_{k}^\top \W_{k} + \V_{k}^\top \V_{k}) - \Tr(\M_{k}^\top \M_{k}). 
\end{align}
Each layer takes as input the previous layer's outputs and a scaled version of the label. 
The model contains a set of $L$ tunable parameters, determining how much we force the neural variables to align with the samples or the labels.

Although we use autograd to compute the gradients, the update rules also admit closed-form solutions as follows
\begin{align}
    \Delta \V_k = \Y \hat{\Z}_{k} , \Delta \W_k = \hat{\Z}_{k-1} \hat{\Z}_{k} , \Delta \M_k = - \hat{\Z}_{k}  \hat{\Z}_{k}^\top.
\end{align}
%
%
The objective Eq.~\eqref{eq:S2M_obj} can also be used for semi-supervised learning, where we can set $\alpha_k$ to 0 for unavailable labels and 1 for available labels, resembling \cite{genkin2019neural}. 
%

Other objective functions, which directly consider the notion of similarities for optimizing NNs as in \cite{pogodin2020kernelized,ma2020hsic}, have been proposed and, as part of the framework, lead to local learning rules or alternatives to BP. 
%
%
These models are nonetheless not greedy. 
Our model is also reminiscent of the work of \citet{nokland2019training}. 

\subsection{Numerical evaluation}

\begin{figure}[!h]
\begin{center}
    \includegraphics[width=.38\textwidth]{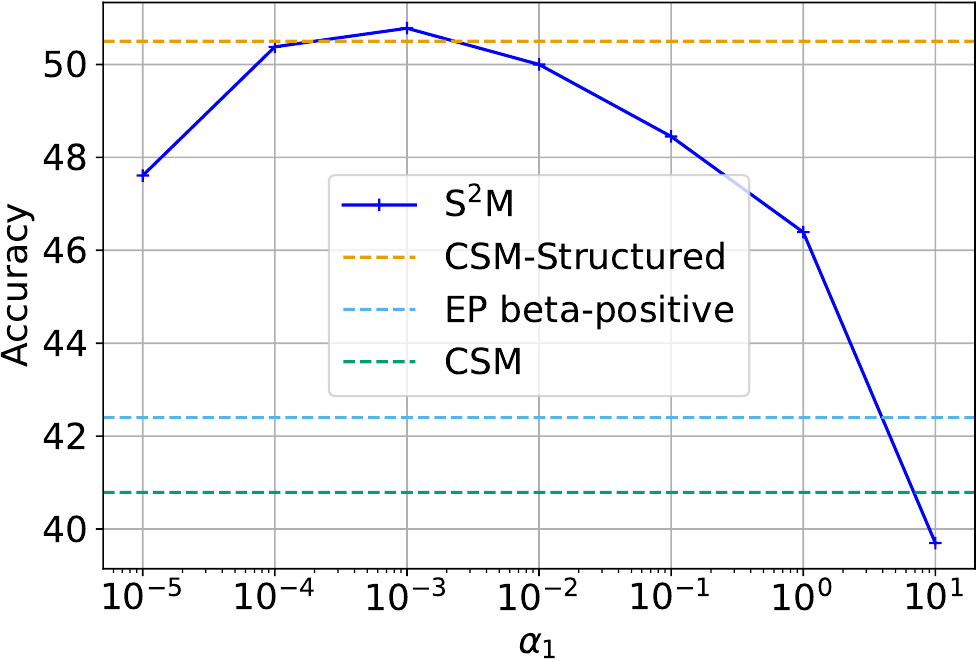}
    \vspace{-0.2cm}
    \caption{Evaluation of S$^2$M. We show classification on CIFAR10 as a function of $\alpha_1$ for a single-layer network of 10 neurons.}
    \label{fig:supervised}
    \end{center}
    \vspace{-0.5cm}
\end{figure}

We evaluated a single layer S$^2$M on CIFAR-10 dataset \cite{krizhevsky2009learning} with different values of $\alpha_1 \in \{10^{-5},\ldots,10^{1}\}$.
The results are shown in Fig.~\ref{fig:supervised}. 
We observe that the accuracy reaches its peak at an optimal $\alpha_1$.
%
%
%
This indicates that the model achieves the highest accuracy at an optimal value of $\alpha_1$ due to a balanced combination of aligning with labels and sample patches crucial for supervised feature learning.
%

We present a comparative analysis of the performance of our novel algorithm on the CIFAR-10 dataset with Contrastive Similarity Matching (CSM) variants \cite{qin2021contrastive} and Equilibrium Propagation (EP) \cite{scellier2017equilibrium}. The evaluation of accuracy is presented in Table \ref{table:validation_error}, where logistic regression is employed for S$^2$M, instead of SVM, to ensure a fair and meaningful comparison among the approaches.

\begin{table}[!h]
\centering
\begin{tabular}{ |c|c|c|c| } 
\hline
 Algorithm  & Accuracy \\ 
 \hline
 CSM-Structured & 50.50\% \\ 
 \hline
 S$^2$M (Logistic Regression) & 49.11\% \\ 
 \hline
 EP beta-positive & 42.40\% \\ 
 \hline
 CSM & 40.79\% \\ 
 \hline
\end{tabular}
\caption{Comparative analysis of the validation accuracy achieved by fully connected networks using Equilibrium Propagation (EP) with positive beta, Contrastive Similarity Matching (CSM), and our novel algorithm on the CIFAR-10 dataset. In all cases, the networks consist of a single hidden layer.}
\vspace{-.3cm}
\label{table:validation_error}
\end{table}

\section{Pre-training}\label{sec:pre-training}

The introduction of pre-training of NNs was a paradigm shift for deep learning. 
Empirical \cite{erhan2010does} and theoretical works such as for sample complexity \cite{tripuraneni2020theory,du2020few} or the out-of-distribution risk \cite{kumar2022fine} tried to understand the mechanisms. 
%

%
Although various candidates for pre-training models exist, we claim that NSM is better suited for the following reasons. 
Unlike sparse coding models, NSM produces ReLU activation instead of soft thresholding \cite{fadili2006sparse, teti2022lcanets}. 
Unlike Non-negative Matrix Factorization (NMF) models \cite{hoyer2004non}, NSM does not enforce the nonnegativity of the weights. 
Unlike autoencoders \cite{goodfellow2016deep}, NSM does not require decoder stages, reducing by half the number of parameters. 

\subsection{Experimental setup}

\textbf{Step 1. Pre-training. } We initialize a single-layer NSM network with the same number of neurons as the filters in the first layer of LeNet.
Subsequently, we train the NSM by executing the dynamics Eq.~\eqref{eq:y_dynamics_supervised} until convergence is achieved.
We then initialize the first layer of LeNet, with the learned weights $\W$ obtained from the NSM, while decoupling $\M$ and $\V$. We initialize the other layers of LeNet randomly.

\textbf{Step 2. Fine-tuning with BP.}
We proceed with supervised fine-tuning of the LeNet layer through BP for all layers.
%
%
%
%
%

%
%
%

\subsection{Numerical evaluation}

\begin{figure}[!h]
\begin{center}
    \includegraphics[width=.4\textwidth]{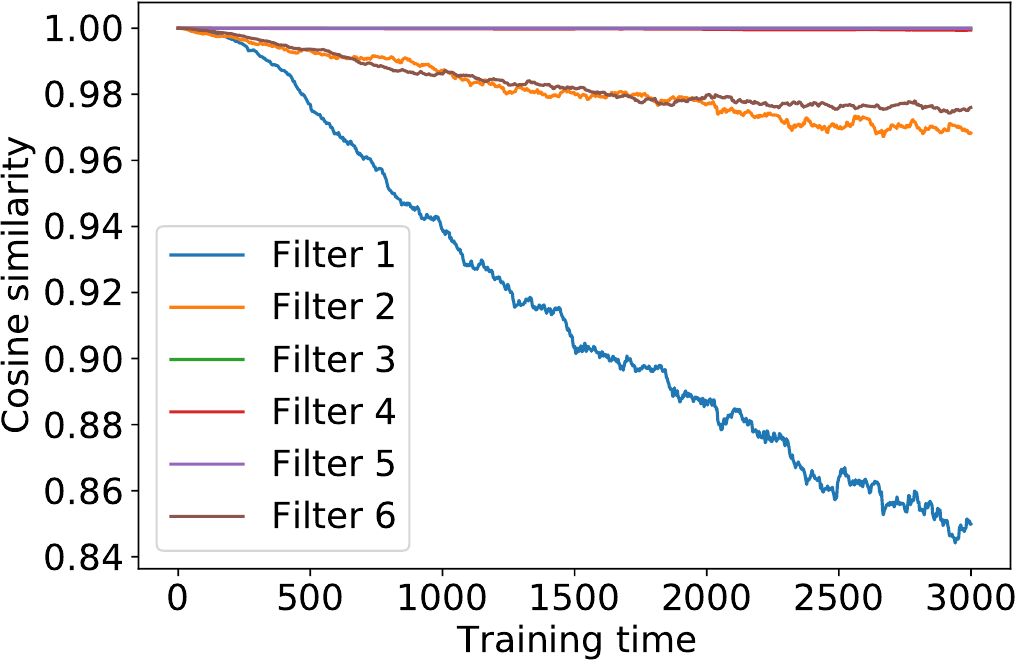}
    \vspace{-0.2cm}
    \caption{Evaluation of unsupervised NSM used for pre-training of LeNet. We measure the cosine similarity between the weights pre-trained with unsupervised NSM and those at each training step of BP. Three of the six pre-trained filters show minimal rotation during BP training.} 
    \label{fig:pretrain_unsupervised}
    \end{center}
    \vspace{-0.5cm}
\end{figure}

\begin{figure}[!h]
\begin{center}
    \includegraphics[width=.4\textwidth]{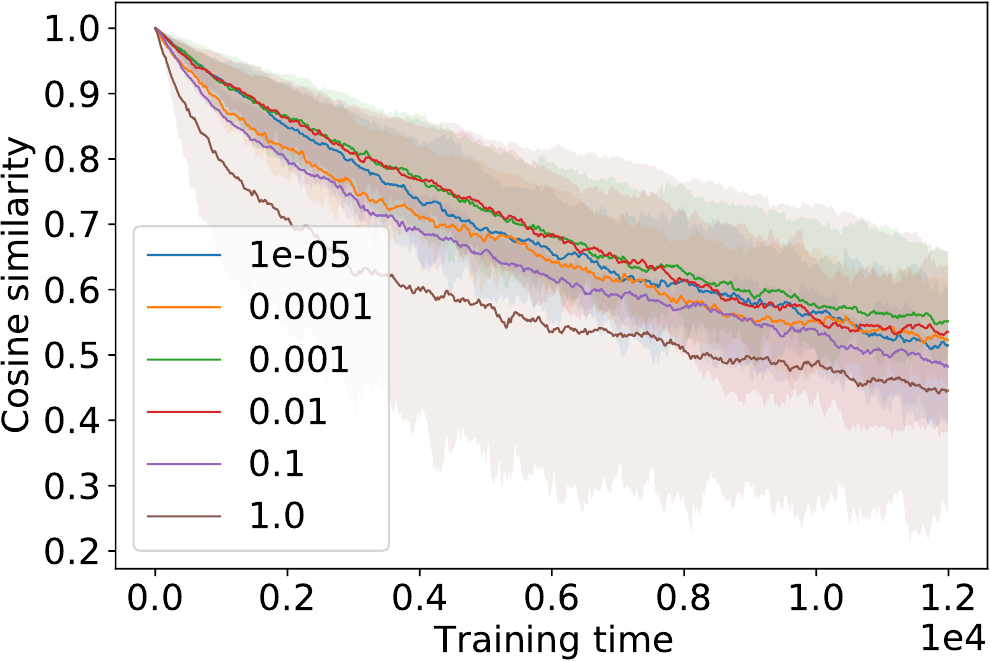}
    \vspace{-0.2cm}
    \caption{Evaluation of supervised NSM used for pre-training of LeNet. We measure the cosine similarity between the weights pre-trained with NSM and those at each training step of BP for different $\alpha_1$. The solid line shows the median value for the six filters averaged over $100$ trials. The shaded region represents the 25 and 75 error percentiles.} 
    \label{fig:pretrain}
    \end{center}
    \vspace{-0.5cm}
\end{figure}

In this section, we present the numerical evaluation of our approach. 
First, we examine the results of unsupervised pre-training of a CNN architecture on the MNIST dataset, as shown in Figure \ref{fig:pretrain_unsupervised}. 
To measure the effectiveness of pre-training, we compute the cosine similarity distance between the pre-trained weights and the evolving filters of the CNN during BP training.
We observe that multiple filters exhibit minimal rotation during BP training, indicating their stability and retention of initial orientations from pre-training. 
This stability suggests that the pre-trained features align well with the final BP-trained features, demonstrating the effective utilization of unsupervised pre-trained knowledge.
%
%
%
%
%
%
Next, in Figure \ref{fig:pretrain}, we analyze the median evolution of filters during BP training for different values of $\alpha_1$, while the shaded regions represent the 25th and 75th percentiles of errors. 
The median rotation of filters during BP training is observed to be lowest for an optimum $\alpha_1$. 
This finding suggests that the filters are most stable and retain their initial orientations from pre-training at the optimum $\alpha_1$.
%
%
%
%
%
%
%

%

In summary, both our unsupervised and supervised pre-training approaches demonstrate the stability of filters and their alignment with the final BP-trained features. These findings provide evidence of the effective utilization of pre-trained knowledge for improving the performance of CNN architectures.



\section{Conclusion and future work}\label{sec:conclusion}

This study represents an initial investigation into the potential of NSM as a localized learning alternative to BP. Our contributions encompass various aspects, including the development of a scalable convolutional NSM implementation using PyTorch, the introduction of a localized supervised objective, and the exploration of NSM-based pre-training for models such as LeNet. These proposed models enhance overall model performance and facilitate efficient learning processes. 
Future work will focus on further exploring and refining the SM framework to uncover its full potential in the realm of localized learning.

%
%
%

\section{Acknowledgement}\label{sec:acknowledgement}

We acknowledge the support of the Simons Foundation,
in particular, the grant SF 626323 to AS, which partly funded this research. We thank our Flatiron Institute and Rutgers University colleagues for their valuable feedback. YB would like to thank Abdul Canatar for insightful discussions. Also, we would like to thank Tiberiu Tesileanu for his assistance in creating the Python package pynsm. The package is available at \url{https://pypi.org/project/pynsm/}. We recommend it to anyone interested in implementing SM-based algorithms in PyTorch. 

\bibliography{example_paper}
\bibliographystyle{icml2023}



\end{document}